\documentclass[10pt,twocolumn,letterpaper]{article}

\usepackage{cvpr}
\usepackage{times}
\usepackage{epsfig}
\usepackage{graphicx}
\usepackage{amsmath}
\usepackage{amssymb}
\usepackage{multirow}
\usepackage{footnote}

\usepackage[pagebackref=true,breaklinks=true,letterpaper=true,colorlinks,bookmarks=false]{hyperref}

\cvprfinalcopy 

\def\cvprPaperID{****} 

\ifcvprfinal\fi

\begin{document}
	\title{Guided Optical Flow Learning}
	
	\author{
		Yi Zhu$^{1}$ \quad Zhenzhong Lan$^{2}$ \quad Shawn Newsam$^{1}$ \quad Alexander G. Hauptmann$^{2}$\\
		$^{1}$University of California, Merced \quad \quad $^{2}$Carnegie Mellon University\\
		{\tt\small \{yzhu25,snewsam\}@ucmerced.edu} \quad \quad {\tt\small \{lanzhzh,alex\}@cs.cmu.edu}
	}
	\maketitle
	
	
	\begin{abstract}
		We study the unsupervised learning of CNNs for optical flow estimation using proxy ground truth data. Supervised CNNs, due to their immense learning capacity, have shown superior performance on a range of computer vision problems including optical flow prediction. 
		They however require the ground truth flow which is usually not accessible except on limited synthetic data. 
		Without the guidance of ground truth optical flow, unsupervised CNNs often perform worse as they are naturally ill-conditioned.
		We therefore propose a novel framework in which proxy ground truth data generated from classical approaches is used to guide the CNN learning. The models are further refined in an unsupervised fashion using an image reconstruction loss. Our guided learning approach is competitive with or superior to state-of-the-art approaches on three standard benchmarks yet is completely unsupervised and can run in real time.
	\end{abstract}

	\section{Introduction}
	\label{sec:intro}
	
	Optical flow contains valuable information for general image sequence analysis due to its capability to represent motion. It is widely used in vision tasks such as human action recognition \cite{twostream2014,depth2action,hidden_zhu_17}, semantic segmentation \cite{videoSeg_cvpr15}, video frame prediction \cite{beyond_mse_iclr16}, video object tracking etc. 
	
	Classical approaches for estimating optical flow are often based on a variational model and solved as an energy minimization process \cite{Horn_Schunck,opticalFlowWarp2004,brox_flow_matching_11}. They remain top performers on a number of evaluation benchmarks; however, most of them are too slow to be used in real time applications. 
	Due to the great success of Convolutional Neural Network (CNN), several works \cite{flownet,sceneflow2016} have proposed using CNNs to estimate the motion between image pairs and have achieved promising results. Although they are much more efficient than classical approaches, these methods require supervision and cannot apply to real world data where the ground truth is not easily accessible. 
	Thus, some recent works \cite{AhmadiICIP2016,jasonUnsup2016,densenet_denseflow_icip17} have investigated unsupervised learning through novel loss functions but they often perform worse than supervised ones. 
	
\begin{figure}[t]
		\centering
		\includegraphics[width=1.0\linewidth,trim=0 0 0 0,clip]{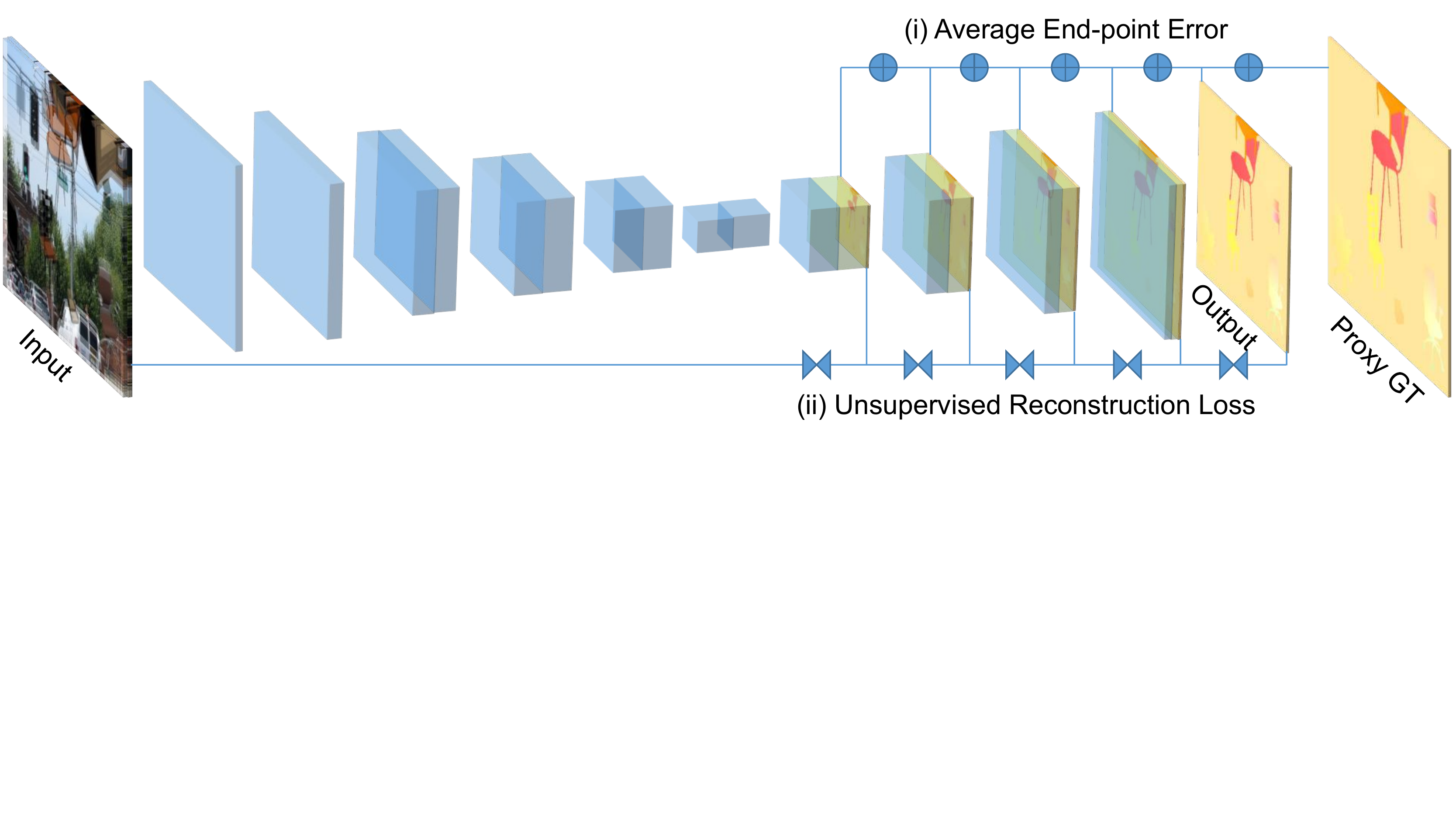}
		\vspace{-16ex}
		\caption{An overview of our proposed guided learning framework. $\oplus$ denotes computing the per-pixel endpoint error with respect to the proxy ground truth flow. $\bowtie$ represents the inverse warping and unsupervised reconstruction loss with respect to the input image pairs. }
		\label{fig:overview}
	\end{figure}

	To improve the accuracy of unsupervised CNNs for optical flow estimation, we propose to use the results of classical methods as guidance for our unsupervised learning process. We refer to this as novel \textit{guided optical flow learning} as shown in Fig. \ref{fig:overview}. 
	Specifically, there are two stages. (i) We generate proxy ground truth flow using classical approaches, and then train a supervised CNN with them. (ii) We fine tune the learned models by minimizing an image reconstruction loss. By training the CNNs using proxy ground truth, we hope to provide a good initialization point for subsequent network learning. By fine tuning the models on target datasets, we hope to overcome the risk that CNN might have learned the failure cases of the classical approaches. The entire learning framework is thus unsupervised.
	
	Our contributions are two-fold. First, we demonstrate that supervised CNNs can learn to estimate optical flow well even when only guided using noisy proxy ground truth data generated from classical methods. Second, we show that fine tuning the learned models for target datasets by minimizing a reconstruction loss further improves performance. Our proposed guided learning is completely unsupervised and achieves competitive or superior performance to state-of-the-art, real time approaches on standard benchmarks. 
	
	
	\section{Method}
	\label{sec:method}
	Given an adjacent frame pair $I_{1}$ and  $I_{2}$, our goal is to learn a model that can estimate the per-pixel motion field $(U, V)$ between the two images accurately and efficiently. $U$ and $V$ are the horizontal and vertical displacements, respectively. We describe our proxy ground truth guided framework in Section \ref{sec:semi}, and the unsupervised fine tuning strategy in Section \ref{sec:unsup}.
	
	\subsection{Proxy Ground Truth Guidance}
	\label{sec:semi}
	Current approaches to the supervised training of CNNs for estimating optical flow use synthetic ground truth datasets. These synthetic motions/scenes are quite different from real ones which limits the generalizability of the learned models. And, even constructing synthetic dataset requires a lot of manual effort \cite{mpi_sintel}. The current largest synthetic datasets with dense ground truth optical flow, Flying Chairs \cite{flownet} and FlyingThings3D \cite{sceneflow2016}, consist of only $22$k image pairs which is not ideal for deep learning especially for such an ill-conditioned problem as motion estimation.  
	In order for CNN-based optical flow estimation to reach its full potential, a learning framework is needed that can scale the size of the training data. Unsupervised learning is one ideal way to achieve this scaling because it does not require ground truth flow.
	
	
	Classical approaches to optical flow estimation are unsupervised in that there is no learning process involved \cite{Horn_Schunck,opticalFlowWarp2004,brox_flow_matching_11,flow_fields_iccv15,patch_match_cvpr16}. They only require the image pairs as input, with some extra assumptions (like image brightness constancy, gradient constancy, smoothness) and information (like motion boundaries, dense image matching). These non-CNN based classical methods currently achieve the best performance on standard benchmarks and are thus considered the state-of-the-art. Inspired by their good performance, we conjecture that \textit{these approaches can be used to generate proxy ground truth data for training CNN-based optical flow estimators}.
	
	In this work, we choose FlowFields \cite{flow_fields_iccv15} as our classical optical flow estimator. 
	To our knowledge, it is one of the most accurate flow estimators among the published work. We hope that by using FlowFields to generate proxy ground truth, we can learn to estimate motion between image pairs as effectively as using the true ground truth.
	
	For fair comparison, we use the ``FlowNet Simple'' network as descried in \cite{flownet} as our supervised CNN architecture. 
	This allows us to compare our guided learning approach to using the true ground truth, particularly with respect to how well the learned models generalize to other datasets. We use endpoint error (EPE) as our guided loss since it is the standard error measure for optical flow evaluation
	\begin{equation}
	L_{\text{epe}} = \frac{1}{N} \sum \sqrt{(U - U^{\prime})^{2} + (V - V^{\prime})^{2}},
	\label{eq:epe_loss}
	\end{equation}
	where $N$ denotes the total number of pixels in $I_{1}$. $U$ and $V$ are the proxy ground truth flow fields while $U^{\prime}$ and $V^{\prime}$ are the flow estimates from the CNN.
	
	\subsection{Unsupervised Fine Tuning}
	\label{sec:unsup}
	As stated in Section \ref{sec:intro}, a potential drawback to using classical approaches to create training data is that the quality of this data will necessarily be limited by the accuracy of the estimator. If a classical approach fails to detect certain motion patterns, a network trained on the proxy ground truth is also likely to miss these patterns. This leads us to ask if there is other unsupervised guidance that can improve the network training? 
	
	The unsupervised approach of \cite{jasonUnsup2016} treats optical flow estimation as an image reconstruction problem based on the intuition that if the estimated flow and the next frame can be used to reconstruct the current frame then the network has learned useful representations of the underlying motions. During training, the loss is computed as the photometric error between the true current frame $I_{1}$ and the inverse-warped next frame $I_{1}^{\prime}$
	\begin{equation}
	L_{\text{reconst}} = \frac{1}{N} \sum_{i, j}^{N} \rho ( I_{1}(i, j) - I_{1}^{\prime}(i,j) ),
	\label{eq:reconstruction_loss}
	\end{equation}
	where $I_{1}^{\prime}(i,j) = I_{2}(i+U_{i,j}, j+V_{i,j})$. The inverse warp is performed using a spatial transformer module \cite{stn_nips15} inside the CNN. We use a robust convex error function, the generalized Charbonnier penalty $\rho(x) = (x^{2} + \epsilon^{2})^{\alpha}$, to reduce the influence of outliers. This reconstruction loss is similar to the brightness constancy objective in classical variational formulations but is quite different from the EPE loss in the proxy ground truth guided learning. We thus propose fine tuning our model using this reconstruction loss as an additional unsupervised guide.
	
	
	During fine tuning, the total energy we aim to minimize is a simple weighted sum of the EPE loss and the image reconstruction loss
	\begin{equation}
	L(U,V; I_{1}, I_{2}) = L_{\text{epe}} + \lambda \cdot L_{\text{reconst}},
	\label{eq:total_loss}
	\end{equation}
	where $\lambda$ controls the level of reconstruction guidance. Note that we could add additional unsupervised guides like a gradient constancy assumption or an edge-aware weighted smoothness loss \cite{monodepthlr2016} to further fine tune our models.
	
	An overview of our guided learning framework with both the proxy ground truth guidance and the unsupervised fine tuning is illustrated in Fig. \ref{fig:overview}.
	
	\section{Experiments}
	\label{sec:experiments}
	
	\subsection{Datasets}
	\label{sec:datasets}
	\textbf{Flying Chairs} \cite{flownet} is a synthetic dataset designed specifically for training CNNs to estimate optical flow. It is created by applying affine transformations to real images and synthetically rendered chairs. The dataset contains 22,872 image pairs: 22,232 training and 640 test samples according to the standard evaluation split. 
	
	\noindent \textbf{MPI Sintel} \cite{mpi_sintel} is also a synthetic dataset derived from a short open source animated 3D movie. There are 1,628 frames, 1,064 for training and 564 for testing. It is the most widely adopted benchmark to compare optical flow estimators. In this work, we only report performance on its final pass because it contains sufficiently realistic scenes including natural image degradations.
	
	\noindent \textbf{KITTI Optical Flow 2012} \cite{Geiger2012CVPR} is a real world dataset collected from a driving platform. It consists of 194 training image pairs and 195 test pairs with sparse ground truth flow. We report the average EPE in total for the test set.
	
	We consider guided learning with and without fine tuning. In the no fine tuning regime, the model is trained using the proxy ground truth produced using a classical estimator. In the fine tuning regime, the model is first trained using the proxy ground truth and then fine tuned using both the proxy ground truth and the reconstruction guide. The Sintel and KITTI datasets are too small to produce enough proxy ground truth to train our model from scratch so the models evaluated on these datasets are first pretrained on the Chairs dataset. These models are then either applied to the Sintel and KITTI datasets without fine tuning or are fine tuned using the target dataset (proxy ground truth).
	
	\subsection{Implementation}
	\label{sec:implementation}
	
	As shown in Fig. \ref{fig:overview}, our architecture consists of contractive and expanding parts. In the no fine tuning learning regime, we calculate the per-pixel EPE loss for each expansion. There are $5$ expansions resulting in $5$ losses. We use the same loss weights as in \cite{flownet}. The models are trained using Adam optimization with the default parameter values $\beta_{1}=0.9$ and $\beta_{2}=0.999$. The initial learning rate is set to $10^{-4}$ and divided by half every $100$k iterations after the first $300$k. We end our training at $600$k iterations.
	
	In the fine tuning learning regime, we calculate both the EPE and reconstruction loss for each expansion. Thus there are a total of $10$ losses. 
	The generalized Charbonnier parameter $\alpha$ is set to $0.25$ in the reconstruction loss. $\lambda$ is $0.1$. We use the default Adam optimization with a fixed learning rate of $10^{-6}$ and training is stopped at $10$k iterations. 
	
	We apply the same intensive data augmentation as in \cite{flownet} to prevent over-fitting in both learning regimes. The proxy ground truth is computed using the FlowFields binary kindly provided by authors in \cite{flow_fields_iccv15}.
	
	
	\begin{table}
		\begin{center}
			\resizebox{\columnwidth}{!}{%
				\begin{tabular}{ | c | c  | c | c |}
					\hline
					Method										&    Chairs    &    Sintel    & KITTI \\
					\hline		
					FlowFields \cite{flow_fields_iccv15}							& 	$2.45$    & $5.81$  & $3.5$\\	
					FlowNetS (Ground Truth) 	\cite{flownet}						&   $2.71$ 	 & $8.43$  & $9.1$\\	
					UnsupFlowNet \cite{jasonUnsup2016}								&   $5.30$ 	&  $11.19$  & $11.3$\\	
					\hline
					FlowNetS (FlowFields)	&   $3.34$ 		 & $8.05$  & $9.7$\\
					FlowNetS (FlowFields) + Unsup	&   $3.01$ 		 & $7.96$  & $9.5$\\
					\hline
				\end{tabular}
			}
		    \vspace{1ex}
			\caption{Results reported using average EPE, lower is better. Bottom section shows our guided learning results, the models are trained using the FlowFields proxy ground truth. The last row includes fine tuning. \label{tab:result1}}
			\vspace{-4ex}
		\end{center}
	\end{table} 
	
	
	\subsection{Results and Discussion}
	\label{sec:discussion}
	
	We have three observations given the results in Table \ref{tab:result1}.
	
	\noindent \textbf{Observation $\mathbf{1}$}: \textit{We can use proxy ground truth generated by state-of-the-art classical flow estimators to train CNNs for optical flow prediction}. A model trained using the FlowFields proxy ground truth achieves an average EPE of $3.34$ on Chairs which is comparable to the $2.71$ achieved by the model trained using the true ground truth. Note that the proxy ground truth is still quite noisy with an average EPE of $2.45$ away from the true ground truth.
	
	\begin{figure*}[t]
		\centering
		\includegraphics[width=0.95\linewidth,trim=0 0 0 0,clip]{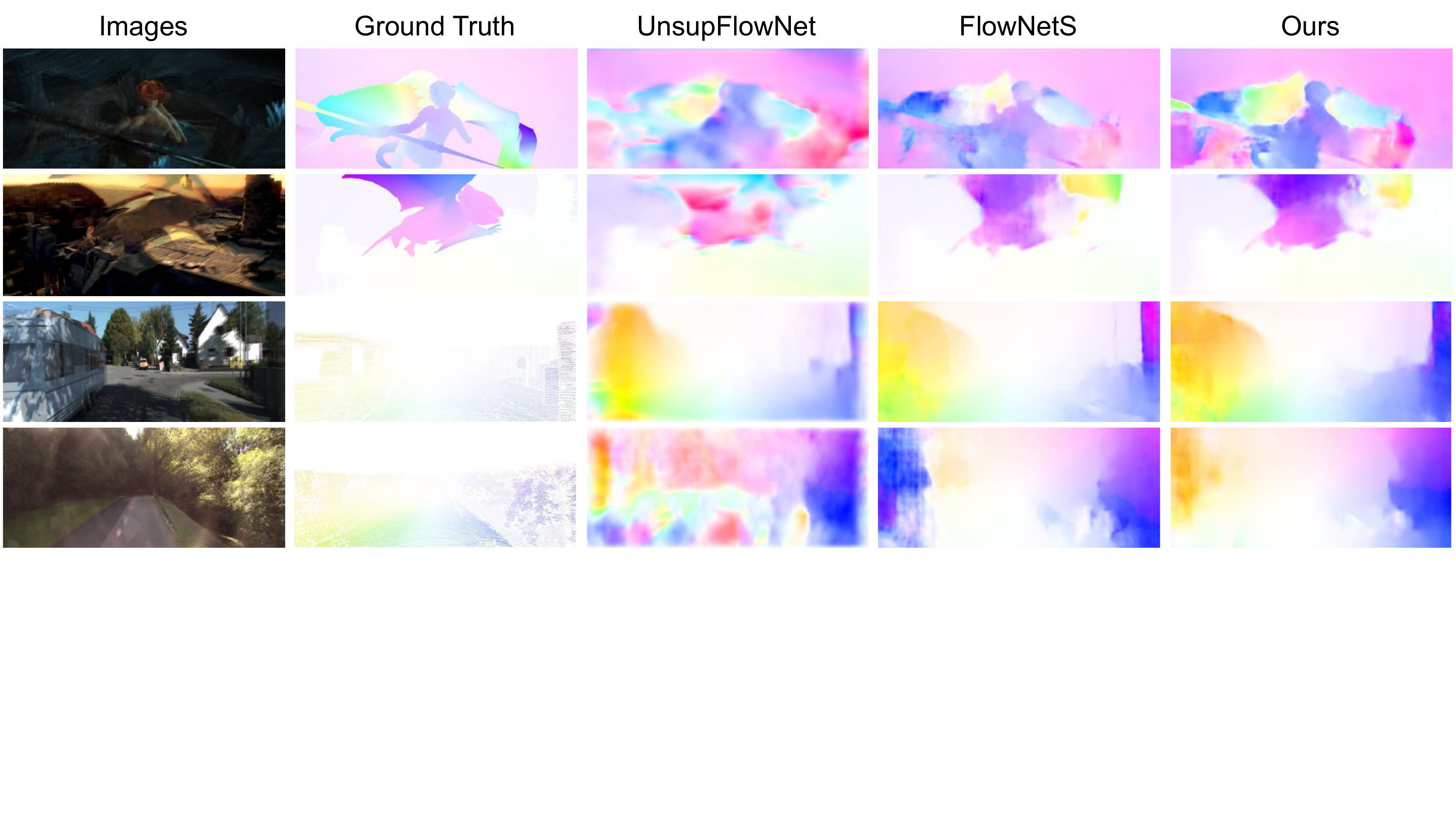}
		\vspace{-18ex}
		\caption{Visual examples of predicted optical flow from different methods. Top two are from Sintel, and bottom two from KITTI.} 
		\label{fig:comparison}
		\vspace{-2ex}
	\end{figure*}
	
	
	The model trained using the FlowFields proxy ground truth (EPE 3.34) performs worse than the FlowFields estimator (EPE 2.45), which is expected. This is because FlowFields adopts a hierarchical approach which is non-local in the image space. It also uses dense correspondence to capture image details. Thus, FlowFields itself can output crisp motion boundaries and accurate flow. However, unlike the CNN model, it cannot run in real time.
	
	\noindent \textbf{Observation $\mathbf{2}$}: \textit{Sometime, training using proxy ground truth can generalize better than training using the true ground truth.} The model trained using the Chairs proxy ground truth (computed with FlowFields) performs better (EPE 8.05) on Sintel than the model trained using the Chairs true ground truth (EPE 8.43). We make similar observations for KITTI\footnote{Note that FlowNetS's performance on KITTI (EPE 9.1) is fine tuned.}. This improved generalization might result from over-fitting when training with the true ground truth since the three datasets are quite different with respect to object and motion types. The proxy is noisier which could serve as a form of data augmentation for unseen motion types.
	
	In addition, we experiment on directly training a Sintel model from scratch without using the pretrained Chairs model. We use the same implementation details. The performance is about one and half pixel worse in terms of EPE than using the pretrained model. Therefore, pretraining CNNs on a large dataset (with either true or proxy ground truth data) is important for optical flow estimation.
	
	
	\noindent \textbf{Observation $\mathbf{3}$}: \textit{Our proposed fine tuning regime improves performance on all three datasets.} Fine tuning results in an average EPE decrease from $3.34$ to $3.01$ for Chairs, $8.05$ to $7.96$ for Sintel, and $9.7$ to $9.5$ for KITTI. Note that an average EPE of $3.01$ for Chairs is very close to performance of the supervised model FlowNetS (EPE $2.71$). This demonstrates that image reconstruction loss is effective as an additional unsupervised guide for motion learning. It can act like fine tuning without requiring ground truth flow of the target dataset. 
	
	
	We also investigate training a network from scratch using a joint training regime. That is, using both $L_{\text{epe}}$ and $L_{\text{reconst}}$, not only using $L_{\text{reconst}}$ in the fine tuning stage. The performance was worse on all three benchmarks. The reason might be that pretraining using just the proxy ground truth prevents the model from becoming trapped in local minima. It thus can provide a good initialization for further network learning. A joint training regime using both losses may hurt the network's convergence in the beginning.
	
	However, we expect unsupervised learning to bring more complementarity. Image reconstruction loss may not be the most appropriate guidance for learning optical flow prediction. We will explore how to best incorporate additional unsupervised objectives in future work.
	
	
	\subsection{Comparison to State-of-the-Art}
	We compare our proposed method to recent state-of-the-art approaches. We only consider approaches that are fast because optical flow is often used in time sensitive applications. 
	We evaluated all CNN-based approaches on a workstation with Intel Core I7 with 4.00GHz and a Nvidia Titan X GPU. For classical approaches, we just use their reported runtime.
	As shown in Table \ref{tab:result2}, our method performs the best for Sintel even though it does not require the true ground truth for training. For Chairs, we achieve on par performance with \cite{flownet}. For KITTI, we perform inferior to \cite{pca_flow_cvpr15}. This is likely because the flow in KITTI is caused purely by the motion of the car so the segmentation into layers performed in \cite{pca_flow_cvpr15} helps in capturing motion boundaries. Our approach outperforms the state-of-the-art unsupervised approaches of \cite{AhmadiICIP2016,jasonUnsup2016} by a large margin, thus demonstrating the effectiveness of our proposed guided learning using proxy ground truth and image reconstruction. Visual comparison of Sintel and KITTI results are shown in Fig. \ref{fig:comparison}. We can see that UnsupFlowNet \cite{jasonUnsup2016} is able to produce reasonable flow fields estimation, but quite noisy. And it doesn't perform well in highly saturated and very dark regions. Our results are much more detailed and smoothed due to the proxy guidance and unsupervised fine tuning.
	
	\section{Conclusion}
	\label{sec:conclusion}
	We propose a guided optical flow learning framework which is unsupervised and results in an estimator that can run in real time. We show that proxy ground truth data produced using state-of-the-art classical estimators can be used to train CNNs. This allows the training sets to scale which is important for deep learning. We also show that training using proxy ground truth can result in better generalization than training using the true ground truth. And, finally, we also show that an unsupervised image reconstruction loss can provide further learning guidance.
	
	More broadly, we introduce a paradigm which can be integrated into future state-of-the-art motion estimation networks \cite{spynet_16} to improve performance. In future work, we plan to experiment with large-scale video corpora to learn non-rigid real world motion patterns rather than just learning limited motions found in synthetic datasets.
	
	\noindent \textbf{Acknowledgements} This work was funded in part by a National Science Foundation CAREER grant, $\#$IIS-1150115. We gratefully acknowledge NVIDIA Corporation through the donation of the Titan X GPU used in this work.
	
	\begin{table}
		\begin{center}
			\resizebox{0.9\columnwidth}{!}{%
				\begin{tabular}{ | c | c | c | c | c | }
					\hline
					Method										&    Chairs      &    Sintel    & KITTI  & Runtime  \\
					\hline
					EPPM 	\cite{EPPM_cvpr14}							&   $-$ 		 & $8.38$    & $9.2$    &  $0.25$\\	
					PCA-Flow 	\cite{pca_flow_cvpr15}						&   $-$ 		 & $8.65$    & $\mathbf{6.2}$    &  $0.19^{\ast}$\\	
					DIS-Fast \cite{dis_fast_eccv16}							&   $-$ 		 & $10.13$   & $14.4$    &  $0.02^{\ast}$\\	
					\hline
					FlowNetS \cite{flownet} 							&   $\mathbf{2.71}$ 		 & $8.43$   & $9.1$    &  $0.06$\\	
					UnsupFlowNet \cite{jasonUnsup2016}								&   $5.30$ 	 & $11.19$   & $11.3$    &  $0.06$\\	
					USCNN 	\cite{AhmadiICIP2016}						&   $-$ 	 & $8.88$   & $-$    &  $-$\\	
					\hline
					Ours	&   $3.01$ 	 & $\mathbf{7.96}$   & $9.5$    &  $0.06$\\
					\hline
				\end{tabular}
			}
			\vspace{2ex}
			\caption{State-of-the-art comparison, runtime is reported in seconds per frame.  Top: Classical approaches. Middle: CNN-based approaches. Bottom: Ours. $^{\ast}$ indicates the algorithm is evaluated using CPU, while the rest are on GPU. \label{tab:result2}}
			\vspace{-4ex}
		\end{center}
	\end{table} 
	
	
	{\small
	\bibliographystyle{ieee}
	\bibliography{egbib_cvpr}
}

\end{document}